\documentclass[sigconf]{acmart}

\usepackage{graphicx}
\usepackage{url}
\usepackage{booktabs} 
\usepackage{multirow} 
\usepackage[table]{xcolor}
\usepackage{balance}
\AtBeginDocument{%
  }

\copyrightyear{2025}
\acmYear{2025}
\setcopyright{acmlicensed}\acmConference[SIGIR-AP 2025]{Proceedings of the 2025 Annual International ACM SIGIR Conference on Research and Development in Information Retrieval in the Asia Pacific Region}{December 7--10, 2025}{Xi'an, China}
\acmBooktitle{Proceedings of the 2025 Annual International ACM SIGIR Conference on Research and Development in Information Retrieval in the Asia Pacific Region (SIGIR-AP 2025), December 7--10, 2025, Xi'an, China}
\acmDOI{10.1145/3767695.3769513}
\acmISBN{979-8-4007-2218-9/2025/12}

\newcommand{\modelname}{\texttt{ScaleFormer}} 

\begin{document}

\title{\modelname: Span Representation Cumulation for Long-Context Transformer}

\author{Jiangshu Du}
\authornote{Work Done Prior to Amazon}
\email{jdu25@uic.edu}
\affiliation{%
  \institution{University of Illinois Chicago}
  \city{Chicago}
  \state{IL}
  \country{USA}
}

\author{Wenpeng Yin}
\email{wenpeng@psu.edu}
\affiliation{%
  \institution{The Pennsylvania State University}
  \city{University Park}
  \state{PA}
  \country{USA}
}

\author{Philip Yu}
\email{psyu@uic.edu}
\affiliation{%
  \institution{University of Illinois Chicago}
  \city{Chicago}
  \state{IL}
  \country{USA}
}

\renewcommand{\shortauthors}{Jiangshu Du, Wenpeng Yin, \& Philip Yu}
\begin{abstract}
The quadratic complexity of standard self-attention severely limits the application of Transformer-based models to long-context tasks. While efficient Transformer variants exist, they often require architectural changes and costly pre-training from scratch. To circumvent this, we propose \modelname~(\textbf{S}pan Representation \textbf{C}umul\textbf{a}tion for \textbf{L}ong-Cont\textbf{e}xt Trans\textbf{former})—a simple and effective plug-and-play framework that adapts off-the-shelf pre-trained encoder-decoder models to process long sequences without requiring architectural modifications. Our approach segments long inputs into overlapping chunks and generates a compressed, context-aware representation for the decoder. The core of our method is a novel, parameter-free fusion mechanism that endows each chunk's representation with structural awareness of its position within the document. It achieves this by enriching each chunk's boundary representations with cumulative context vectors from all preceding and succeeding chunks. This strategy provides the model with a strong signal of the document's narrative flow, achieves linear complexity, and enables pre-trained models to reason effectively over long-form text. 
Experiments on long-document summarization show that our method is highly competitive with and often outperforms state-of-the-art approaches without requiring architectural modifications or external retrieval mechanisms.
\end{abstract}

\ccsdesc[500]{Computing methodologies~Natural language generation}
\ccsdesc[500]{Computing methodologies}
\ccsdesc[500]{Computing methodologies~Artificial intelligence}
\ccsdesc[500]{Computing methodologies~Natural language processing}

\keywords{Long-context transformers,
Pre-trained language models,
Overlapping segmentation,
Sliding window attention}


\maketitle
\section{Introduction}

The advent of Transformer-based pre-trained language models (PLMs) has marked a paradigm shift in Natural Language Processing (NLP) \cite{vaswani2017attention}. However, the Transformer's foundational self-attention mechanism scales quadratically with input sequence length, creating a substantial barrier to its application in domains that rely on long-form text, such as summarizing scientific articles, analyzing legal documents, or processing entire books \cite{beltagy2020longformer, zaheer2020bigbird, bertsch2023unlimiformer}.

To address this long-context barrier, researchers have developed numerous efficient Transformer variants, often based on sparse attention patterns that reduce complexity from quadratic to linear or near-linear. While effective, these models typically introduce custom architectural changes that require expensive pre-training from scratch. As a result, a critical research direction has emerged: adapting existing powerful models for long-context tasks. This raises a central research question: \textit{How can we adapt existing PLMs to handle long-form text without incurring the prohibitive cost of pre-training new models from scratch?}

The landscape of solutions is diverse. One major line of work involves fundamental \textbf{architectural modifications} to create efficient Transformers, using techniques like sparse attention \cite{beltagy2020longformer, zaheer2020bigbird} or recurrence \cite{dai2019transformerxl}. Another line focuses on \textbf{hierarchical pipelines} that first extract salient information before processing it \cite{pilault2020extractive}. A third—and most relevant to our work—centers on \textbf{plug-and-play frameworks} designed to wrap around existing PLMs \cite{ivgi2023efficient, gunther2025latechunkingcontextualchunk, shang2025longrope2nearlosslessllmcontext, chen2025corecontextawaretransformers}. Prominent examples include SLED \cite{ivgi2023efficient}, which employs a "fusion-in-decoder" strategy \cite{izacard2021leveraging} by concatenating all encoded chunks, and Unlimiformer \cite{bertsch2023unlimiformer}, which augments the model with a retrieval mechanism that accesses relevant parts of the long input from a k-nearest neighbor (kNN) index.

While effective, these plug-and-play methods place the full burden of discerning the document’s overall structure and narrative flow on the decoder, which must infer these relationships from a long sequence of concatenated chunk representations. In this work, we introduce \modelname~(Span Representation Cumulation for Long-Context
Transformer)—a novel framework that offers an alternative approach by explicitly encoding structural information into the input representations before they reach the decoder.

Our key innovation lies in an efficient and parameter-free fusion scheme. Instead of passing all encoded tokens from every segment to the decoder, we represent each segment using only its ``boundaries''—the hidden states of its leftmost and rightmost tokens. These boundary representations are then dynamically enriched through our novel span representation cumulation mechanism. This technique provides each segment with positional awareness by combining its left boundary with a summary of all preceding text and its right boundary with a summary of all succeeding text. This approach achieves linear complexity while enabling off-the-shelf PLMs to perform effectively on long-context tasks.

Our primary contributions are as follows:
\begin{enumerate}
    \item We propose \modelname, a novel and efficient framework for applying pre-trained encoder-decoder models to long sequences with linear complexity, without requiring any modifications to the model's internal architecture.
    \item We introduce a directional boundary fusion mechanism that provides each segment with cumulative, position-aware context, explicitly encoding the document's structure into each chunk's representation.
    \item We conduct extensive experiments on long-document summarization tasks, demonstrating that our method achieves highly competitive performance against strong baselines.
\end{enumerate}

\section{Related Work}

Transformer models have been adapted in various ways to handle long sequences by reducing quadratic self-attention costs. Sparse-attention approaches like Longformer \cite{beltagy2020longformer} and BigBird \cite{zaheer2020bigbird} limit each token’s attention to local windows or selected global tokens. Other efficient attention approximations include Reformer's hashing-based method \cite{kitaev2020reformer}, Routing Transformers’ clustering \cite{roy2020routing}, and low-rank or kernel methods such as Linformer \cite{wang2020linformer} and Performer \cite{choromanski2020performer}.  More recent work continues to push these boundaries, with novel architectures like Infini-attention proposing methods to scale to theoretically infinite contexts by integrating a compressive memory directly into the attention mechanism \cite{munkhdalai2024leave}. These models often require architecture changes and expensive pre-training. In contrast, our method leverages existing pre-trained Transformers without modifying their internal attention mechanism, enabling the efficient reuse of these powerful models.

Retrieval-based methods like RAG \cite{lewis2020rag} and REALM \cite{guu2020realm} retrieve relevant passages before processing them. The Fusion-in-Decoder (FiD) \cite{izacard2021leveraging} strategy processes passages independently and fuses the encoded representations in the decoder, inspiring various approaches in QA and summarization. Recent chunk-based methods, notably SLED \cite{ivgi2023efficient}, segment long inputs and encode chunks separately before decoding. Another parallel line of work adapts decoder-only LLMs using similar chunking principles; for instance, CEPE \cite{yen2024longcontext} also processes long inputs in parallel chunks but requires inserting new cross-attention modules into a frozen decoder to integrate the context. While our method shares SLED’s philosophy of leveraging pre-trained Transformers through chunking, we differ fundamentally in how chunk representations are managed. Instead of using entire chunk encodings, we select only boundary tokens to create compressed representations that are then infused with explicit directional context, a feature absent in prior work.

Another related approach, Unlimiformer \cite{bertsch2023unlimiformer}, extends input lengths by offloading decoder attention to an external kNN index. This allows for extremely long contexts but adds complexity. In contrast, our model operates within the standard Transformer framework without additional indexing, offering a simpler solution. Overall, our method balances efficiency and simplicity by avoiding architectural modifications, extensive retraining, and external retrieval structures.

\section{The \modelname~ Framework}
We propose \modelname, a simple yet effective framework for applying existing PLMs to long-sequence tasks. The core idea is to segment long inputs into overlapping chunks and create a compact, fused representation for each segment that captures both its local content and its structural position within the overall document.

\begin{figure}[h]
    \centering
    \includegraphics[width=\linewidth]{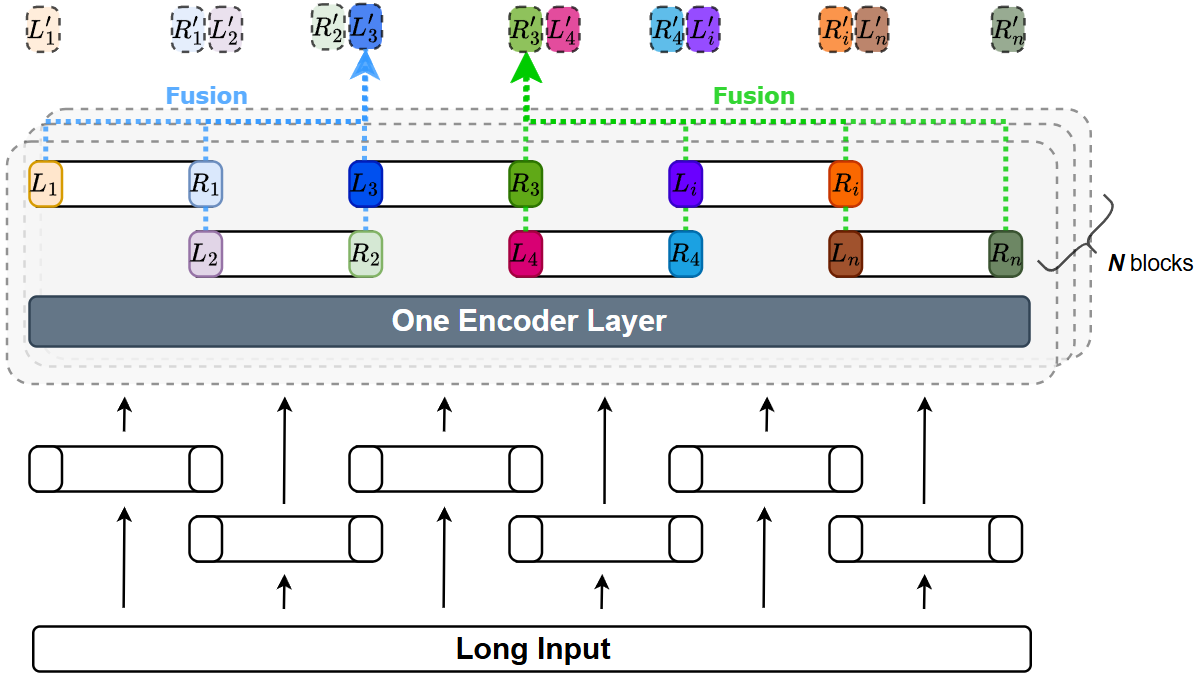} 
    \caption{Overview of the \modelname~framework. A long input document is segmented into overlapping chunks, each encoded independently. Boundary token representations (Left and Right) are extracted and fused with directional context. The Left boundary of a chunk is fused with context from prior chunks, and the Right boundary is fused with context from subsequent chunks, providing structural awareness.}
    \label{fig:DCFusion_arch}
\end{figure}

As illustrated in Figure \ref{fig:DCFusion_arch}, the \modelname~framework proceeds in several steps. First, a long input document is tokenized and partitioned into multiple overlapping segments. Second, each chunk is processed independently by the encoder. Third, from each sequence of hidden states, we extract the representations corresponding to the first few and last few tokens—the Left (L) and Right (R) boundaries. Fourth, these boundary vectors undergo our novel span representation cumulation step. Finally, the resulting fused representations from all chunks are concatenated and passed to the decoder.

\subsection{Overlapping Segment Encoding}
Formally, let an input sequence of $N$ tokens be denoted as $X = \{x_1, x_2, \dots, x_N\}$. We partition $X$ into $C$ overlapping segments $S_1, \allowbreak S_2, \allowbreak \dots, \allowbreak S_C$. Each segment $S_i$ has a fixed length $L$, which fits within the model's maximum context window. The overlap between consecutive segments $S_i$ and $S_{i+1}$ is a fixed number of tokens, $O$.

Each segment $S_i$ is then encoded independently using the pre-trained encoder, $M_{\text{enc}}$, to produce a sequence of top-layer hidden states $H_i \in \mathbb{R}^{L \times d}$, where $d$ is the hidden dimension of the model:
\begin{equation}
    H_i = M_{\text{enc}}(S_i)
\end{equation}
From each $H_i$, we extract the left boundary $L_i \in \mathbb{R}^{k \times d}$ (first $k$ tokens) and the right boundary $R_i \in \mathbb{R}^{k \times d}$ (last $k$ tokens). This process is inherently parallelizable and scales linearly with the input length $N$.

\subsection{Span Representation Cumulation}
After the parallel encoding of segments, the primary challenge is to present this information to the decoder in a way that preserves the document's global structure. A naive approach of simply concatenating the encoded chunks (e.g., $[H_1, H_2, ..., H_C]$) poses a significant challenge for the decoder. This method presents it with a long, structurally undifferentiated sequence of hidden states where the original relationships between chunks are lost. 

Consequently, the decoder's cross-attention mechanism is burdened with a difficult dual task: it must first attempt to infer the document's original structure and narrative flow from this flat sequence, and only then can it identify and synthesize the most salient information for summarization. This inefficient use of the model's representational capacity can lead to errors in the final output, such as a lack of coherence, incorrect temporal ordering, or a failure to capture a main thesis developed across the entire document.

Our \modelname~ is designed specifically to alleviate this burden. Instead of forcing the decoder to infer structure, we explicitly inject it into each segment's representation before the decoding stage. By enriching each chunk's boundaries with cumulative information about its preceding and succeeding context, we provide explicit signals about its position and role, allowing the decoder to focus its resources on generating a high-quality, globally-aware summary.

\textbf{Boundary Extraction:} For each encoded segment $H_i$, we extract its boundary representations. These are the hidden states corresponding to the first $k$ tokens, which we call the left boundary $L_i \in \mathbb{R}^{k \times d}$, and the last $k$ tokens, the right boundary $R_i \in \mathbb{R}^{k \times d}$. These boundaries serve as proxies for the contextualized information at the edges of each segment, which are crucial for linking information across chunks.

\textbf{Directional Context:} For each segment $S_i$, we compute a backward-looking context vector, $\text{ctx}^{\text{back}}_i$, and a forward-looking context vector, $\text{ctx}^{\text{fwd}}_i$.

The \textbf{backward context} for segment $i$ is the average of its own left boundary $L_i$ and all boundary vectors ($L_j$ and $R_j$) from preceding segments ($j < i$):
\begin{equation}
    \text{ctx}^{\text{back}}_i = \frac{1}{2i-1} \left( L_i + \sum_{j=1}^{i-1} (L_j + R_j) \right) \quad \text{for } i > 1
\end{equation}
The \textbf{forward context} for segment $i$ is the average of its own right boundary $R_i$ and all boundary vectors from succeeding segments ($j > i$):
\begin{equation}
    \text{ctx}^{\text{fwd}}_i = \frac{1}{2(C-i)+1} \left( R_i + \sum_{j=i+1}^{C} (L_j + R_j) \right) \quad \text{for } i < C
\end{equation}
Here, $C$ represents the total number of chunks. For the first chunk's backward context and the last chunk's forward context, the context vector is simply the local boundary itself.

\textbf{Weighted Fusion:} The final fused representations, $L'_i$ and $R'_i$, are created by blending the local boundary vector with its corresponding directional context vector, controlled by a hyperparameter $\alpha \in [0, 1]$.
\begin{equation}
    L'_i = \alpha \cdot L_i + (1 - \alpha) \cdot \text{ctx}^{\text{back}}_i
\end{equation}
\begin{equation}
    R'_i = \alpha \cdot R_i + (1 - \alpha) \cdot \text{ctx}^{\text{fwd}}_i
\end{equation}
\subsection{Middle Token Sampling}
While boundary vectors are effective for capturing inter-segment context, they may not fully represent the unique content within each segment. To enhance the local representation, we also sample $m$ "middle" tokens, $M_i \in \mathbb{R}^{m \times d}$, from the interior of each encoded chunk $H_i$. The purpose of these middle tokens is to provide the decoder with direct, unaltered snapshots of the segment's core content. This strategy further improves performance by providing richer local information while maintaining a highly compressed overall representation. This sampling can be seen as a form of text compression, conceptually similar to recent work in prompt compression. For instance, LLMLingua \cite{jiang2023llmlingua, llmlingua2} compresses prompts by using a small language model to identify and remove less informative tokens. Although our current implementation employs random sampling, we plan to explore more effective token-selection methods in future work.

\subsection{Final Decoder Input}
The final sequence of hidden states passed to the decoder $M_{\text{dec}}$ is the concatenation of the fused boundary representations and the sampled middle tokens from all segments:
\begin{equation}
    H_{\text{final}} = \text{concat}(L'_1, M_1, R'_1, L'_2, M_2, R'_2, \dots, L'_C, M_C, R'_C)
\end{equation}
The length of this sequence is $C \times (2k + m)$, which is substantially smaller than the $C \times L$ tokens used by a naive fusion-in-decoder approach like SLED.

\subsection{Complexity Analysis}
The computational complexity of our \modelname~framework scales favorably with the input sequence length $N$. The encoding step involves $C$ independent forward passes through the encoder on sequences of length $L$. Since $C \approx N / (L-O)$, the total complexity for encoding is proportional to $(N/L) \times O(L^2) = O(N \cdot L)$. As $L$ is a fixed constant (the model's maximum context size), the encoding complexity is linear with respect to the input length, i.e., $O(N)$. The directional context calculation involves a single pass over the $C$ chunk boundaries, which is computationally negligible. The decoder's complexity is dominated by the length of its input, $C \times (2k + m)$. Since $C$ is proportional to $N$, and $k, m$ are small constants, the effective complexity of our entire framework is \textbf{linear} with respect to the input sequence length $N$.

\begin{table*}[h!]
\centering
\caption{BART-base summarization results on dev and test sets. R-1/2/L are ROUGE scores; BS is BERT-Score. Best scores in each column are in \textbf{bold}. For tied scores, both are bolded.}
\label{tab:bart_results}
\begin{tabular}{l|l|cccc|cccc|cccc}
\toprule
& \multirow{2}{*}{\textbf{Method}} & \multicolumn{4}{c|}{\textbf{SummScreen}} & \multicolumn{4}{c|}{\textbf{GovReport}} & \multicolumn{4}{c}{\textbf{BookSum}} \\
& & R-1 & R-2 & R-L & BS & R-1 & R-2 & R-L & BS & R-1 & R-2 & R-L & BS \\
\midrule
\multirow{6}{*}{\rotatebox[origin=c]{90}{\textbf{Dev}}}
& Standard (Trunc.) & 30.0 & 6.5 & 17.7 & 56.7 & 47.7 & 18.5 & 22.3 & 64.0 & 35.3 & 8.8 & 13.9 & 70.2 \\
& SLED& 34.2 & 8.2 & 19.2 & 58.8 & 55.5 & 24.8 & 25.8 & 66.9 & - & - & - & - \\
& Mem. Trans. & 32.8 & 7.6 & 19.3 & 57.7 & 55.8 & 25.6 & 26.9 & 67.7 & - & - & - & - \\
& Unlimiformer & 35.0 & 8.3 & 19.6 & 58.4 & \textbf{57.4} & \textbf{26.4} & 27.9 & 68.2 & 35.6 & \textbf{11.1} & 15.0 & 71.1 \\
& \cellcolor{gray!10}\modelname~(Ours) & \cellcolor{gray!10}34.3 & \cellcolor{gray!10}8.1 & \cellcolor{gray!10}19.3 & \cellcolor{gray!10}58.6 & \cellcolor{gray!10}56.1 & \cellcolor{gray!10}25.2 & \cellcolor{gray!10}27.5 & \cellcolor{gray!10}67.6 & \cellcolor{gray!10}35.7 & \cellcolor{gray!10}9.7 & \cellcolor{gray!10}14.7 & \cellcolor{gray!10}70.9 \\
& \cellcolor{gray!10}\modelname~+ Middle & \cellcolor{gray!10}\textbf{34.7} & \cellcolor{gray!10}\textbf{9.2} & \cellcolor{gray!10}\textbf{20.1} & \cellcolor{gray!10}\textbf{58.9} & \cellcolor{gray!10}\textbf{57.4} & \cellcolor{gray!10}26.0 & \cellcolor{gray!10}\textbf{28.1} & \cellcolor{gray!10}\textbf{68.3} & \cellcolor{gray!10}\textbf{36.7} & \cellcolor{gray!10}10.9 & \cellcolor{gray!10}\textbf{15.8} & \cellcolor{gray!10}\textbf{71.5} \\
\midrule
\multirow{6}{*}{\rotatebox[origin=c]{90}{\textbf{Test}}}
& Standard (Trunc.)  & 29.7 & 6.2 & 17.7 & 56.3 & 48.7 & 19.2 & 22.8 & 64.3 & 36.4 & 7.6 & 15.3 & 70.5 \\
& SLED & 32.7 & 7.9 & 19.1 & 58.4 & 54.7 & 24.4 & 25.4 & 67.0 & - & - & - & - \\
& Mem. Trans. & 32.7 & 7.4 & 19.2 & 57.4 & 55.2 & 25.1 & 26.4 & 67.5 & 35.6 & 6.4 & 14.6 & 70.9 \\
& Unlimiformer & \textbf{34.7} & 8.5 & 19.9 & 58.5 & 56.6 & \textbf{26.3} & 27.6& 68.2& 37.3 & 6.7 & 15.2 & 71.3 \\
& \cellcolor{gray!10}\modelname~(Ours) & \cellcolor{gray!10}33.3 & \cellcolor{gray!10}8.3 & \cellcolor{gray!10}19.2 & \cellcolor{gray!10}58.7 & \cellcolor{gray!10}56.0 & \cellcolor{gray!10}25.1 & \cellcolor{gray!10}26.8 & \cellcolor{gray!10}67.5 & \cellcolor{gray!10}37.1 & \cellcolor{gray!10}7.1 & \cellcolor{gray!10}15.7 & \cellcolor{gray!10}71.3 \\
& \cellcolor{gray!10}\modelname~+ Middle & \cellcolor{gray!10}34.2 & \cellcolor{gray!10}\textbf{9.0} & \cellcolor{gray!10}\textbf{20.3} & \cellcolor{gray!10}\textbf{59.0} & \cellcolor{gray!10}\textbf{57.0} & \cellcolor{gray!10}25.7 & \cellcolor{gray!10}\textbf{27.7} & \cellcolor{gray!10}\textbf{68.4} & \cellcolor{gray!10}\textbf{39.2} & \cellcolor{gray!10}\textbf{9.3} & \cellcolor{gray!10}\textbf{16.3} & \cellcolor{gray!10}\textbf{71.9} \\
\bottomrule
\end{tabular}
\end{table*}

\section{Experiments}
\subsection{Experimental Setup}
\textbf{Tasks and Datasets:} We evaluate our framework on three long-document summarization benchmarks:
\begin{itemize}
    \item \textbf{SummScreen} \cite{chen2022summscreen}: Comprises TV show transcripts and their summaries, with an average input length of approximately 9,000 tokens.
    \item \textbf{GovReport} \cite{huang2021efficient}: A dataset of lengthy government reports, where inputs are truncated to 16,384 tokens for evaluation consistency.
    \item \textbf{BookSum} \cite{kryscinski2021booksum}: A challenging dataset for narrative summarization of books, with extremely long documents averaging over 140,000 tokens.
\end{itemize}

\textbf{Backbone Models and Baselines:} To demonstrate its generalizability, we apply our method to \textbf{BART-base} and \textbf{T5-base}. We compare against a suite of strong baselines representing different approaches to long-context processing:
\begin{itemize}
    \item \textbf{Standard (Truncation):} This baseline truncates the input to the model's maximum context window (1024 tokens). It serves as a lower bound, showing performance without any long-context adaptation.
    \item \textbf{SLED} \cite{ivgi2023efficient}: A chunk-based method that encodes overlapping segments and concatenates all resulting hidden states for the decoder. It serves as our most direct baseline as it also uses a "fusion-in-decoder" strategy, but crucially, it results in a much longer input sequence for the decoder and lacks an explicit mechanism for encoding document structure.
    \item \textbf{Memorizing Transformers} \cite{wu2022memorizing}: Enhances the Transformer with an approximate k-nearest-neighbor (kNN) memory, allowing attention layers to access more context than fits in the standard window.
    \item \textbf{Unlimiformer} \cite{bertsch2023unlimiformer}: A powerful retrieval-based method that offloads the decoder's cross-attention to a kNN index built from all encoder hidden states.
\end{itemize}

\textbf{Evaluation Metrics:} Following prior work, we report performance using ROUGE-1, ROUGE-2, and ROUGE-L, and BERT-Score on both development and test datasets.

\textbf{Implementation Details:} Our \modelname~method uses a chunk size ($L$) of 1024 and an overlap ($O$) of 150 tokens. We use $k=1$ for boundary tokens. The fusion ratio $\alpha$ was set to 0.5 based on performance on the dev set. We evaluate two variants:
\begin{itemize}
    \item \textbf{\modelname}: Our core model, which uses only the directional fusion mechanism on boundary tokens. The decoder receives a compressed sequence of fused left and right boundary representations from each chunk.
    \item \textbf{\modelname~+ Middle}: Our full model, which augments the fused boundary tokens with $m=300$ randomly sampled "middle" tokens from each chunk's interior. This variant is designed to balance global structural awareness with richer local content.
\end{itemize}

\subsection{Main Results}

We present our findings on both development and test sets, comparing our two variants—\modelname~(fusion only) and \modelname~+ Middle—against all baselines.

\textbf{BART-base Results:} As shown in Table \ref{tab:bart_results}, \modelname~framework demonstrates strong performance. The base \modelname~ outperforms SLED on the test sets of SummScreen and GovReport. On the GovReport test set, for instance, \modelname~achieves a ROUGE-1 score of 56.0, surpassing SLED's 54.7. The addition of middle token sampling (\modelname~+ Middle) provides a substantial boost, establishing our method as state-of-the-art on nearly all benchmarks. On the extremely long-context BookSum dataset, \modelname~+ Middle achieves a ROUGE-1 of 39.2 on the test set, decisively outperforming all baselines including Unlimiformer (37.3), and sets new state-of-the-art scores across all four metrics. On the GovReport test set, it achieves the highest ROUGE-1 (57.0), ROUGE-L (27.7), and BERT-Score (68.4).

\textbf{T5-base Results:} To further validate our framework, we applied it to T5-base. The results, presented in Table \ref{tab:t5_results}, confirm the consistency of our approach. The performance trends mirror the BART-base experiments: \modelname~is a clear improvement over the standard truncation baseline and outperforms SLED. Once again, \modelname~+ Middle outperforms both the standard truncation baseline and SLED across all metrics on the test set.

\begin{table}[h!]
\centering
\caption{T5-base summarization results on dev and test sets. Best scores are in \textbf{bold}.}
\label{tab:t5_results}
\begin{tabular}{l|l|ccc|ccc}
\toprule
& \multirow{2}{*}{\textbf{Method}} & \multicolumn{3}{c|}{\textbf{SummScreen}} & \multicolumn{3}{c}{\textbf{GovReport}} \\
& & R-1 & R-2 & R-L & R-1 & R-2 & R-L \\
\midrule
\multirow{4}{*}{\rotatebox[origin=c]{90}{\textbf{Dev}}}
& Standard (Trunc.)  & 22.2 & 3.7 & 15.3 & 32.8 & 11.7 & 20.2 \\
& SLED & 25.3 & 5.0 & 16.6 & 47.0 & 20.2 & 25.2 \\
& \cellcolor{gray!10}ScaleF. (Ours) & \cellcolor{gray!10}25.5 & \cellcolor{gray!10}5.1 & \cellcolor{gray!10}16.9 & \cellcolor{gray!10}47.9 & \cellcolor{gray!10}20.8 & \cellcolor{gray!10}26.4 \\
& \cellcolor{gray!10}ScaleF. + Middle & \cellcolor{gray!10}\textbf{26.0} & \cellcolor{gray!10}\textbf{5.9} & \cellcolor{gray!10}\textbf{17.3} & \cellcolor{gray!10}\textbf{48.7} & \cellcolor{gray!10}\textbf{21.4} & \cellcolor{gray!10}\textbf{27.7} \\
\midrule
\multirow{4}{*}{\rotatebox[origin=c]{90}{\textbf{Test}}}
& Standard (Trunc.)  & 21.4 & 3.6 & 15.0 & 33.2 & 12.1 & 20.4 \\
& SLED & 24.5 & 4.6 & 16.5 & 46.6 & 20.1 & 25.1 \\
& \cellcolor{gray!10}ScaleF.(Ours) & \cellcolor{gray!10}24.7 & \cellcolor{gray!10}4.7 & \cellcolor{gray!10}16.7 & \cellcolor{gray!10}47.0 & \cellcolor{gray!10}20.3 & \cellcolor{gray!10}25.4 \\
& \cellcolor{gray!10}ScaleF. + Middle & \cellcolor{gray!10}\textbf{25.4} & \cellcolor{gray!10}\textbf{5.1} & \cellcolor{gray!10}\textbf{17.1} & \cellcolor{gray!10}\textbf{47.9} & \cellcolor{gray!10}\textbf{21.2} & \cellcolor{gray!10}\textbf{27.4} \\
\bottomrule
\end{tabular}
\end{table}

\begin{figure*}[t!]
    \centering
    \includegraphics[width=\textwidth]{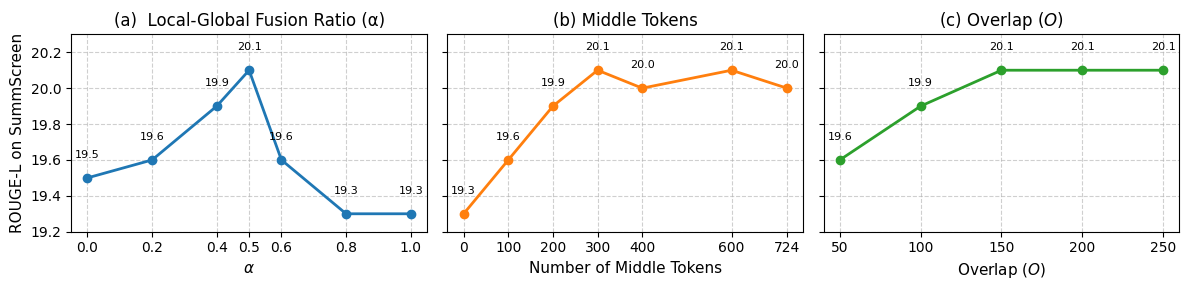} 
    \caption{Ablation study results on the SummScreen dev set. We analyze the impact of (a) the directional fusion ratio $\alpha$, (b) the number of middle tokens sampled ($m$), and (c) the chunk overlap size ($O$). Performance is measured in ROUGE-L.}
    \label{fig:ablation}
\end{figure*}

\section{Ablation and Analysis}
\label{sec:ablation}
To understand the contribution of each component in our framework and analyze its sensitivity, we conduct ablation studies on the SummScreen development set using BART-base. Unless specified otherwise, hyperparameters are kept at their default values ($L=1024$, $O=150$, $m=300$, $\alpha=0.5$). We report ROUGE-L, with results presented in Figure~\ref{fig:ablation}.

\subsection{Effect of the Directional Fusion Ratio ($\alpha$)}
The core of our method is the weighted fusion between local boundaries and their directional context, controlled by $\alpha$. To validate its effectiveness, we analyze performance while varying $\alpha$ from $0.0$ to $1.0$.
\begin{itemize}
    \item An $\alpha$ of \textbf{1.0} represents a \textbf{local-only} model where boundaries are not fused with any directional context, relying only on local information.
    \item An $\alpha$ of \textbf{0.0} represents a \textbf{context-only} model where local boundary information is replaced entirely by the directional average.
\end{itemize}
As shown in Figure~\ref{fig:ablation}(a), the model performs poorly at both extremes. A purely local-only model ($\alpha=1.0$) and a near-context-only model ($\alpha \approx 0.0$) yield low ROUGE-L scores. Performance steadily increases as local and directional contexts are mixed, peaking at $\alpha=0.5$ with a ROUGE-L of 20.1. This confirms our central hypothesis that a balance between segment-specific details and cumulative document context is critical for strong performance.

\subsection{Impact of Middle Token Sampling}
We analyze the utility of including $m$ "middle" tokens from each chunk. Figure~\ref{fig:ablation}(b) shows a clear benefit. The model without any middle tokens ($m=0$) starts at a ROUGE-L of 19.3. Performance sharply increases as more local content is provided, reaching a peak of 20.1 at $m=300$. Adding more tokens beyond this point does not yield further improvements, indicating that 300 tokens provide a sufficient snapshot of local content. This justifies our choice of $m=300$ for our primary model.

\subsection{Analysis of Chunk Overlap}
The overlap size ($O$) between consecutive chunks is critical for ensuring smooth context propagation between segments. In Figure~\ref{fig:ablation}(c), performance improves as the overlap increases from 50 to 150 tokens, with the ROUGE-L score rising from 19.6 to 20.1. Beyond this point, increasing the overlap provides no additional benefit while increasing computational overhead. We therefore use $O=150$ as it achieves the best performance-cost trade-off.

\section{Conclusion}
In this paper, we introduced \modelname, a simple, effective, and parameter-free framework for adapting pre-trained encoder-decoder models to long-context tasks. By segmenting long inputs and fusing chunk boundary representations with directional context, our method provides the model with an explicit understanding of the document's structure, a critical feature often lost in standard chunking approaches. Our method achieves linear complexity without any architectural changes to the underlying model. Extensive experiments on long-document summarization demonstrate that our proposed \modelname, especially when augmented with middle token sampling, achieves state-of-the-art or highly competitive results across multiple benchmarks and backbone models. 

\begin{acks}
The authors appreciate the reviewers for their insightful comments and suggestions.
This work is supported in part by NSF under grants III-2106758, and POSE-2346158
\end{acks}
\bibliographystyle{ACM-Reference-Format}
\balance 
\bibliography{sample-sigconf-authordraft}

\appendix

\end{document}